\documentclass{article}
\usepackage[utf8]{inputenc}

\usepackage{graphicx}
\usepackage{xcolor}

\usepackage{booktabs}
\usepackage{multirow}
\usepackage{authblk}
\usepackage[normalem]{ulem}
\usepackage{topiclongtable}
\useunder{\uline}{\ul}{}

\usepackage[
backend=biber,
style=numeric,
sorting=nyt
]{biblatex}
\title{LLMs for Multi-Modal Knowledge Extraction and Analysis in Intelligence/Safety-Critical Applications}
\author{Brett Israelsen}
\author{Soumalya Sarkar}
\affil{RTX Technology Research Center (RTRC)}
\date{September 2023}

\begin{document}

\maketitle

\begin{abstract}
    Large Language Models have seen rapid progress in capability in recent years; this progress has been accelerating and their capabilities, measured by various benchmarks, are beginning to approach those of humans. There is a strong demand to use such models in a wide variety of applications but, due to unresolved vulnerabilities and limitations, great care needs to be used before applying them to intelligence and safety-critical applications. This paper reviews recent literature related to LLM assessment and vulnerabilities to synthesize the current research landscape and to help understand what advances are most critical to enable use of of these technologies in intelligence and safety-critical applications. The vulnerabilities are broken down into ten high-level categories and overlaid onto a high-level life cycle of an LLM. Some general categories of mitigations are reviewed.
\end{abstract}

\section{Introduction}
With the advent of GPT-3 there seems to have been a rapid acceleration in the observed capabilities of large language models (LLMs) being released. This explosion of capability came as somewhat of a surprise because the main `innovation' was to increase the number of parameters being used in the models (which now number in the hundreds of billions~\cite{Wei2022-zd}). This resulted in `emergent' capabilities, or capabilities that were not explicitly sought during the design and training process~\cite{Wei2022-zd}.

This increase in capability has captured the interest and imaginations of the public and technologists alike, and propelled LLMs into the limelight. While the extent to which LLMs will be integrated into our lives is yet to be seen, it is clear that they will have a profound impact.

One crucial roadblock is that LLMs are still quite limited in their capabilities to perform as expected. These limitations and vulnerabilities are not new, they had been identified before the most recent generation of highly-capable LLMs, but began to receive more attention as scientists and researchers began considering how LLMs might be applied to various safety-critical applications. In such applications there is very little tolerance for behaviors that cannot be predicted and fully understood; safety-critical applications also require that vulnerabilities be addressed before they can be deployed into real-world environments where they can be subject to non-ideal circumstances, and even adversarial attacks. Examples of some use cases that we consider to qualify as `safety-critical' include: (i) Automated requirement generation for software, (ii) Digital assistant for planning, (iii) reasoning and decision making, (iv) multi-modal inspection, (v) interactive ISR with Intelligence analysis, (vi) Contextualized summarization for intelligence analysis, and (vii) survey analytics.

This paper reviews recent LLM literature on the vulnerabilities and limitations of LLMs in order to identify the current understanding, and what approaches are being taken to address them. We further discuss implications to intelligence and safety-critical applications.

\section{LLM Vulnerabilities}
There has been a recent explosion in literature focusing on vulnerabilities of LLMs due to the surprising increase in capability shown initially by GPT and its subsequent versions as well as other similar models released by companies like Google, Facebook, Anthropic and others. This interest isn't new, but has definitely increased in recent months.

In the context of LLMs we define \emph{vulnerabilities} as: properties or behaviors of LLMs (generally, or individually) that make them prone to degraded performance through attack, misuse, or normal operation.

A review of some of the recent literature was performed with the goal of better understanding and synthesizing current understanding about the vulnerabilities of LLMs; the key results of this review are included in Section~\ref{sec:vulnerabilities}. Sections~\ref{sec:models} and \ref{sec:datasets} offer a quick overview of the models and datasets used in the reviewed papers.

\subsection{Literature Synthesis}\label{sec:vulnerabilities}
\begin{figure}[h]
    \centering
    \includegraphics[width=0.95\textwidth]{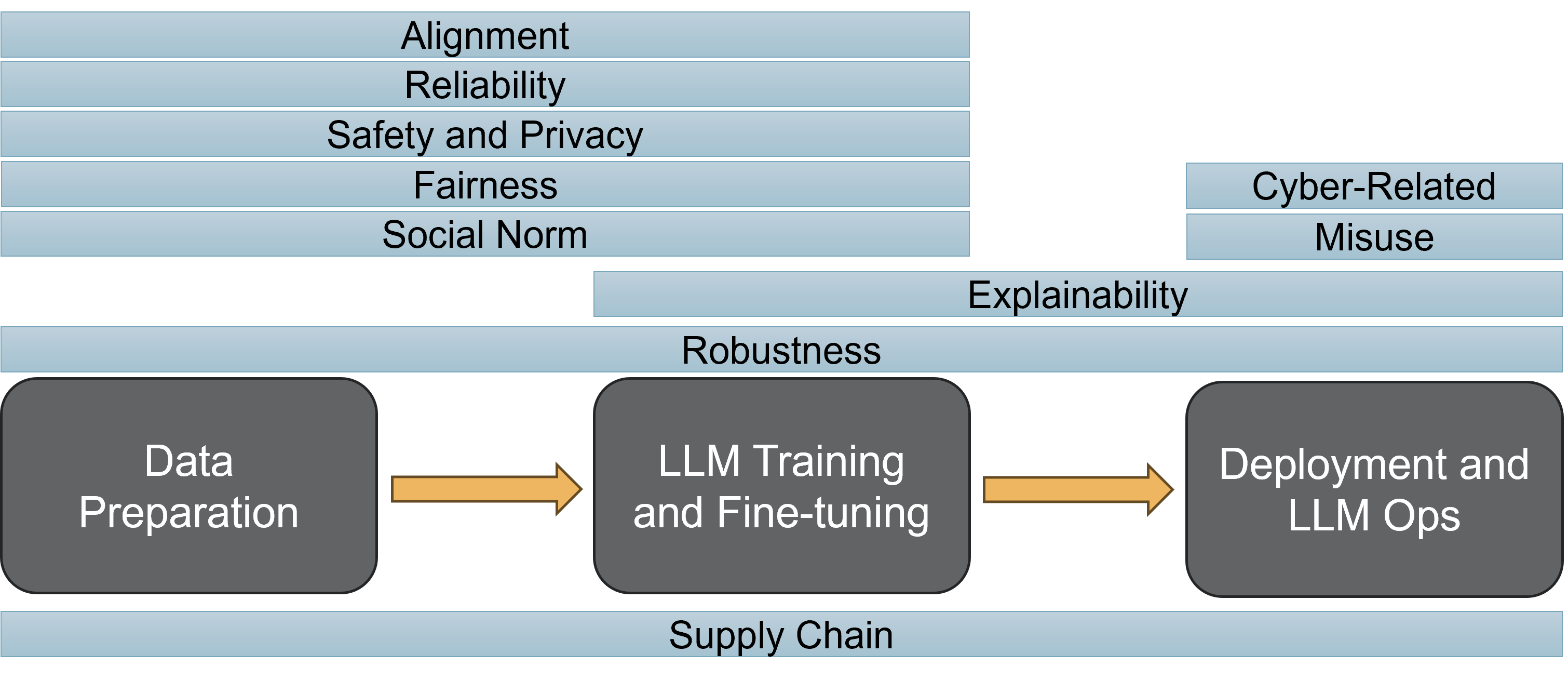}
    \caption{Illustration of LLM vulnerabilities. The vulnerabilities have been overlaid on a high-level life cycle of an LLM to illustrate phases of development where vulnerabilities might be addressed.}
    \label{fig:vulnerabilities}
\end{figure}

There are \emph{many} different lists and assessments of the limitations and vulnerabilities of LLMs that have been produced. The amount of literature referencing such topics has only increased recently. The main goal of this review was to synthesize some of the existing literature to get a better idea of the `landscape'. Based on review of approximately 20 recent LLM papers, Figure~\ref{fig:vulnerabilities} gives an illustration of some high-level classes of vulnerabilities and the stages of the LLM life cycle where they are typically manifest. 

Each of these categories are made up of smaller sub-categories that are listed in Table~\ref{tab:vulnerabilities}, and individual references are given for each of the sub-categories. Short definitions of each vulnerability class follow.

\begin{description}
    \item[Alignment] The degree to which the underlying behavior (driven by the training objectives, loss and reward functions) of the model aligns (or matches) the behavior expected/desired by stake-holders. This category also includes ethical considerations since these are often assumed (wrongly) to be met.
    \item[Reliability] The degree to which the model is capable of performing in a reliable manner. Typical issues that can affect reliability include: `hallucination', `miscalibration', and `misinformation' (outputs that are incorrect).
    \item[Safety and Privacy] This category encompasses behaviors concerned with critical undesirable behaviors. Having a model exhibit such behaviors, even rarely, is not acceptable. Well-known behaviors in this category include: `emergence' (un-programmed and un-tested behaviors), `information leaking' (divulging of sensitive or private information), and `legality' (model providing outputs that support illegal, or otherwise forbidden, behaviors).
    \item[Fairness] Includes model behaviors that are concerned with consistently reliable and balanced treatment across different subgroups including prompts (i.e. different languages and dialects) and responses (i.e. not referring to certain relevant information more often than other relevant information). This class includes behaviors such as: `bias', `injustice', and `performance disparity' (better performance in some contexts than others).
    \item[Social Norm] Since these systems are interacting with individuals who live in a society, it is important that they accurately reflect the values of that society. This category is related to behaviors such as `cultural insensitivity', `toxicity', and `unawareness of emotions'. Such behaviors are at the center of interpersonal relationships; those who violate such norms typically pay a social price, and deployed LLM models should comply as well.
    \item[Cyber-Related] At the most basic level, LLMs share the same vulnerabilities of all information-technology systems. Cybersecurity vulnerabilities are fairly well understood, but have unique ways in which they can be manifest in the LLM life cycle. Some typical vulnerabilities included in this class include `prompt injection' (inserting certain text into prompts that override LLM controls), `insecure output handling' (giving the LLM excessive, and unsafe, amounts of freedom in outputs), and `model denial of service' (ability of third-parties to deny service to others through overwhelming the model servers with requests).
    \item[Misuse] This class of vulnerabilities is based on the ways in which LLMs can be misused (on purpose or not). Some examples of misuse include `copyright infringements' (where the LLM produces content that infringes on copyright), `propaganda' (where an LLM might be utilized to produce highly believable, but misleading, information), and `social engineering' (when LLM outputs might be used to impersonate real people in order to try to manipulate them).
    \item[Robustness] This class includes vulnerabilities related to circumstances that may cause an LLM to perform differently in certain situations than in others. This includes various considerations about training data (including poisoning and prompt attacks, dynamic data such as real-time news, and distributions shifts), this can also include possible performance changes observed when utilizing `personas' (asking the LLM to act as if it were a certain individual, or had certain expertise). 
    \item[Explainability] This category is concerned with the ability of human users to understand the reasoning process of the LLM. When an LLM is asked questions that require reasoning it is important to know the limitations of the LLM to reason. Also, more generally, in many critical circumstances it is crucial for those with decision-making authority to be able to check and verify certain information before proceeding.
    \item[Supply Chain] As LLMs become more complex it is increasingly common that different pieces are sourced from third-parties. This results in the overall performance being reliant on the cumulative performance of many disparate parts in the supply chain. This means quality-control and standards are necessary to ensure all sub-components (and their inevitable updates) can be tested for quality. Also, as `foundation models' become more common, there will likely be many derivative systems that heavily rely on them possibly exposing them to any vulnerability of the foundation model.
\end{description} 

Each class of vulnerabilities is contained within a light-blue box; the width of the box varies to overlap with the stage of the LLM life cycle where the vulnerability is most likely to be manifest. It is important to note that the assignment to life cycle stages is not exact, but should help highlight `typical' areas of concern. As an example, the `Robustness' category spans all three phases of the LLM life cycle because robustness vulnerabilities are manifest in each of these phases. In the `Data Preparation' stage: poisoning, and data curation; in the `Training' stage: distribution shifts, and (again) data curation; and in the `Deployment' stage: prompt attacks, and distribution shifts. In contrast the `Misuse' category sits only in the `Deployment' stage because that is the only time in which the LLM is actually used (issuing responses for prompts).

\TopicSetWidth{=}

\begin{topiclongtable}{@{}Fp{0.2\textwidth} Tp{0.5\textwidth} Tp{0.3\textwidth}@{}}
\caption{Vulnerabilities broken down by category}
\label{tab:vulnerabilities}
\\

\textbf{Category} & \textbf{Vulnerabilities}                  & \textbf{Papers} \\
\hline\endfirsthead
\multicolumn{3}{c}%
{{\bfseries Table \thetable\ continued from previous page}} \\
\textbf{Category} & Vulnerabilities                  & Papers                                                                                                                                                                                                          \\
\hline\endhead
\TopicLine \Topic[alignment]         & alignment                        & \cite{Askell2021-mf}, \cite{Kumar2023-hl}, \cite{Li2023-hf}, \cite{Shevlane2023-rp}, \cite{Wolf2023-nq}                                                                                                             \\
\TopicLine \Topic[alignment]         & ethics                           & \cite{Bommasani2021-ir}                                                                                                                                                                                         \\
\TopicLine \Topic[cyber]             & excessive agency                 & \cite{OWASP_Foundation_undated-qz}                                                                                                                                                                              \\
\TopicLine \Topic[cyber]             & insecure output handling         & \cite{OWASP_Foundation_undated-qz}                                                                                                                                                                              \\
\TopicLine \Topic[cyber]             & insecure plugin design           & \cite{OWASP_Foundation_undated-qz}                                                                                                                                                                              \\
\TopicLine \Topic[cyber]             & model denial of service          & \cite{OWASP_Foundation_undated-qz}                                                                                                                                                                              \\
\TopicLine \Topic[cyber]             & model theft                      & \cite{OWASP_Foundation_undated-qz}                                                                                                                                                                              \\
\TopicLine \Topic[cyber]           & prompt injection                 & \cite{Kumar2023-hl}, \cite{OWASP_Foundation_undated-qz}, \cite{Zou2023-it}                                                                                                                                        \\
\TopicLine \Topic[explainability]    & causal reasoning                 & \cite{Liu2023-yx}                                                                                                                                                                                               \\
\TopicLine \Topic[explainability]    & interpretability                 & \cite{Liu2023-yx}                                                                                                                                                                                               \\
\TopicLine \Topic[explainability]    & logical reasoning                & \cite{Liu2023-yx}                                                                                                                                                                                               \\
\TopicLine \Topic[fairness]          & bias                             & \cite{Bender2021-wc}, \cite{Liu2023-yx}, \cite{Rutinowski2023-kk}, \cite{Steinhardt_undated-vi}                                                                                                                    \\
\TopicLine \Topic[fairness]          & inequity                         & \cite{Bommasani2021-ir}                                                                                                                                                                                         \\
\TopicLine \Topic[fairness]          & injustice                        & \cite{Liu2023-yx}                                                                                                                                                                                               \\
\TopicLine \Topic[fairness]          & lack of diversity                & \cite{Bender2021-wc}                                                                                                                                                                                            \\
\TopicLine \Topic[fairness]          & performance disparity            & \cite{Liu2023-yx}                                                                                                                                                                                               \\
\TopicLine \Topic[fairness]          & preference bias                  & \cite{Liu2023-yx}                                                                                                                                                                                               \\
\TopicLine \Topic[fairness]          & stereotyping                     & \cite{Liu2023-yx}                                                                                                                                                                                               \\
\TopicLine \Topic[misuse]            & copyright                        & \cite{Liu2023-yx}                                                                                                                                                                                               \\
\TopicLine \Topic[misuse]            & cyberattack                      & \cite{Liu2023-yx}                                                                                                                                                                                               \\
\TopicLine \Topic[misuse]            & mistrust                         & \cite{Weidinger2021-hs}                                                                                                                                                                                         \\
\TopicLine \Topic[misuse]            & misuse                           & \cite{Bommasani2021-ir}, \cite{Christ2023-bn}, \cite{Weidinger2021-hs}                                                                                                                                            \\
\TopicLine \Topic[misuse]            & overreliance                     & \cite{OWASP_Foundation_undated-qz}                                                                                                                                                                              \\
\TopicLine \Topic[misuse]            & propaganda                       & \cite{Liu2023-yx}                                                                                                                                                                                               \\
\TopicLine \Topic[misuse]            & social-engineering               & \cite{Liu2023-yx}                                                                                                                                                                                               \\
\TopicLine \Topic[reliability]       & hallucination                    & \cite{Liu2023-yx}, \cite{Manakul2023-yk}, \cite{Reese2023-yb}                                                                                                                                                     \\
\TopicLine \Topic[reliability]       & inconsistency                    & \cite{Liu2023-yx}                                                                                                                                                                                               \\
\TopicLine \Topic[reliability]       & miscalibration                   & \cite{Amayuelas2023-aa}, \cite{Angelopoulos2021-uw}, \cite{Duan2023-zr}, \cite{Huang2023-dq}, \cite{Kuhn2023-yi}, \cite{Kumar2023-ji}, \cite{Lin2023-dn}, \cite{Liu2023-yx}, \cite{Shevlane2023-rp}, \cite{Xiong2023-ve} \\
\TopicLine \Topic[reliability]       & misinformation                   & \cite{Liu2023-yx}, \cite{Weidinger2021-hs}                                                                                                                                                                       \\
\TopicLine \Topic[reliability]       & sychopancy                       & \cite{Liu2023-yx}                                                                                                                                                                                               \\
\TopicLine \Topic[reliability]       & unfaithful explanations          & \cite{Turpin2023-kl}                                                                                                                                                                                            \\
\TopicLine \Topic[robustness]        & data curation                    & \cite{Bender2021-wc}                                                                                                                                                                                            \\
\TopicLine \Topic[robustness]        & distribution shifts              & \cite{Liu2023-yx}                                                                                                                                                                                               \\
\TopicLine \Topic[robustness]        & dynamic data                     & \cite{Bender2021-wc}                                                                                                                                                                                            \\
\TopicLine \Topic[robustness]        & imitating personas               & \cite{Wolf2023-nq}                                                                                                                                                                                              \\
\TopicLine \Topic[robustness]        & interventional effect            & \cite{Liu2023-yx}                                                                                                                                                                                               \\
\TopicLine \Topic[robustness]        & paradigm shifts                  & \cite{Liu2023-yx}                                                                                                                                                                                               \\
\TopicLine \Topic[robustness]        & poisoning attacks                & \cite{Liu2023-yx}                                                                                                                                                                                               \\
\TopicLine \Topic[robustness]        & prompt attacks                   & \cite{Liu2023-yx}                                                                                                                                                                                               \\
\TopicLine \Topic[robustness]        & training data poisoning          & \cite{OWASP_Foundation_undated-qz}                                                                                                                                                                              \\
\TopicLine \Topic[safety/privacy]    & adult content                    & \cite{Liu2023-yx}                                                                                                                                                                                               \\
\TopicLine \Topic[safety/privacy]    & emergence                        & \cite{Ganguli2022-zr}, \cite{Steinhardt_undated-vi}, \cite{Wei2022-zd}                                                                                                                                            \\
\TopicLine \Topic[safety/privacy]    & information leaking              & \cite{Weidinger2021-hs}                                                                                                                                                                                         \\
\TopicLine \Topic[safety/privacy]    & legality                         & \cite{Bommasani2021-ir}                                                                                                                                                                                         \\
\TopicLine \Topic[safety/privacy]    & mental health                    & \cite{Liu2023-yx}                                                                                                                                                                                               \\
\TopicLine \Topic[safety/privacy]    & privacy violation                & \cite{Liu2023-yx}                                                                                                                                                                                               \\
\TopicLine \Topic[safety/privacy]    & sensitive information disclosure & \cite{OWASP_Foundation_undated-qz}                                                                                                                                                                              \\
\TopicLine \Topic[safety/privacy]    & unlawful conduct                 & \cite{Liu2023-yx}                                                                                                                                                                                               \\
\TopicLine \Topic[safety/privacy]    & violence                         & \cite{Liu2023-yx}                                                                                                                                                                                               \\
\TopicLine \Topic[social norm]       & cultural insensitivity           & \cite{Liu2023-yx}                                                                                                                                                                                               \\
\TopicLine \Topic[social norm]       & toxicity                         & \cite{Liu2023-yx}, \cite{Steinhardt_undated-vi}                                                                                                                                                                  \\
\TopicLine \Topic[social norm]       & unawareness of emotions          & \cite{Liu2023-yx}                                                                                                                                                                                               \\
\TopicLine \Topic[supply chain]      & economic incentives              & \cite{Bommasani2021-ir}                                                                                                                                                                                         \\
\TopicLine \Topic[supply chain]      & environment                      & \cite{Bommasani2021-ir}                                                                                                                                                                                         \\
\TopicLine \Topic[supply chain]      & homogenization                   & \cite{Steinhardt_undated-vi}                                                                                                                                                                                    \\
\TopicLine \Topic[supply chain]      & lack of industry standards       & \cite{Wei2022-zd}                                                                                                                                                                                               \\
\TopicLine \Topic[supply chain]      & supply chain vulnerabilities     & \cite{OWASP_Foundation_undated-qz}                                                                                                                                                                             
\end{topiclongtable}

\subsection{Referenced Models}\label{sec:models}
In the set of papers we reviewed there were thirty-two models that were discussed. These models are listed in Table~\ref{tab:models}\footnote{One `model' listed in Table~\ref{tab:models} is called `general'. This represents papers that focused on topics that are general to all LLMs and not specific to particular implementations and/or architectures.} with a short description and references to the particular papers that specifically mentioned them. These models represent a reasonably large subset of the state-of-the-art at the time of writing of this paper (Fall 2023).

Some of the models represented are `fine-tuned' versions of other models (i.e. `alpaca' is fine-tuned from `LLaMA', and `deberta' is extended from `RoBERTa'. Some of the models (like `llama' and `opt') are open source, and are able to be analyzed in more detail than other proprietary models (i.e. the `gpt' family, and `claude'). Judging by number of references in the surveyed papers, the `gpt' models and `llama' are the most popular.

\begin{longtable}{@{}lp{0.6\textwidth}p{0.2\textwidth}@{}}
\caption{Summary of LLM models included in surveyed papers}
\label{tab:models}\\
\toprule
\textbf{Model}       & \textbf{Description}                                           & \textbf{Papers}                                                                                                                                                                                                                                                   \\* \midrule
\endfirsthead
\multicolumn{3}{c}%
{{\bfseries Table \thetable\ continued from previous page}} \\
\toprule
\textbf{Model}       & \textbf{Description}                                           & \textbf{Papers}                                                                                                                                                                                                                                                   \\* \midrule
\endhead
\bottomrule
\endfoot
\endlastfoot
\textbf{alpaca}      & fine-tuned from LLaMA                                          & \cite{Amayuelas2023-aa}, \cite{Choi2023-hb}, \cite{Li2023-hf}                                                                                                                                                                                                     \\
\textbf{bert}        & family of language models introduced by Google in 2018         & \cite{Choi2023-hb}                                                                                                                                                                                                                                                \\
\textbf{bloom}       & fine-tuned from Megatron-LM                                    & \cite{Choi2023-hb}                                                                                                                                                                                                                                                \\
\textbf{chatglm}     & open bilingual language model based on GLM                     & \cite{Zou2023-it}                                                                                                                                                                                                                                                 \\
\textbf{chatgpt}     & OpenAI proprietary model                                       & \cite{Rutinowski2023-kk}                                                                                                                                                                                                                                          \\
\textbf{claude}      & Anthropic proprietary model                                    & \cite{Turpin2023-kl}, \cite{Wu2023-my}, \cite{Zou2023-it}                                                                                                                                                                                                         \\
\textbf{codegen}     & autoregressive LM for program synthesis                        & \cite{Huang2023-dq}                                                                                                                                                                                                                                               \\
\textbf{custom}      & decoder-only transformer model                                 & \cite{Askell2021-mf}                                                                                                                                                                                                                                              \\
\textbf{davinci-002} & GPT-3.5 variant using instruction tuning                       & \cite{Duan2023-zr}                                                                                                                                                                                                                                                \\
\textbf{davinci-003} & GPT-3.5 variant using RLHF                                     & \cite{Amayuelas2023-aa}, \cite{Duan2023-zr}                                                                                                                                                                                                                       \\
\textbf{deberta}     & extension of RoBERTa                                           & \cite{Choi2023-hb}                                                                                                                                                                                                                                                \\
\textbf{falcon}      & open model released by TII                                     & \cite{Zou2023-it}                                                                                                                                                                                                                                                 \\
\textbf{flan}        & Google model with instruction tuning                           & \cite{Choi2023-hb}                                                                                                                                                                                                                                                \\
\textbf{general}     & not specific to particular architecture                        & \cite{Angelopoulos2021-uw}, \cite{Bender2021-wc},   \cite{Bommasani2021-ir}, \cite{Christ2023-bn},   \cite{OWASP_Foundation_undated-qz}, \cite{Shevlane2023-rp},   \cite{Steinhardt_undated-vi}, \cite{Wei2022-zd}, \cite{Weidinger2021-hs},   \cite{Wolf2023-nq} \\
\textbf{google lm}   & collection of Google LMs                                       & \cite{Ganguli2022-zr}                                                                                                                                                                                                                                             \\
\textbf{gopher}      & 280 Billion parameter model from Google                        & \cite{Ganguli2022-zr}                                                                                                                                                                                                                                             \\
\textbf{gpt-2}       & Open AI; 2019 release                                          & \cite{Choi2023-hb}, \cite{Huang2023-dq}                                                                                                                                                                                                                           \\
\textbf{gpt-3}       & Open AI; 2020 release, instruction tuning and training on code & \cite{Amayuelas2023-aa}, \cite{Ganguli2022-zr}, \cite{Huang2023-dq},   \cite{Lin2023-dn}, \cite{Manakul2023-yk}, \cite{Turpin2023-kl},   \cite{Wu2023-my}, \cite{Xiong2023-ve}, \cite{Zou2023-it}                                                                 \\
\textbf{gpt-3.5}     & Open AI; 2022 release, supervised instruction tuning and RLHF                & \cite{Amayuelas2023-aa}, \cite{Huang2023-dq}, \cite{Lin2023-dn},   \cite{Turpin2023-kl}, \cite{Wu2023-my}, \cite{Xiong2023-ve},   \cite{Zou2023-it}                                                                                                               \\
\textbf{gpt-4}       & Open AI; 2023 release, increased memory, includes multi-modal inputs         & \cite{Amayuelas2023-aa}, \cite{Reese2023-yb}, \cite{Wu2023-my},   \cite{Xiong2023-ve}, \cite{Zou2023-it}                                                                                                                                                          \\
\textbf{gpt-j}       & Eleuther AI                                                    & \cite{Choi2023-hb}                                                                                                                                                                                                                                                \\
\textbf{gpt-neo}     & Eleuther AI                                                    & \cite{Li2023-hf}                                                                                                                                                                                                                                                  \\
\textbf{guanaco}     & based off of LLaMA model family; used LoRA fine-tuning         & \cite{Zou2023-it}                                                                                                                                                                                                                                                 \\
\textbf{incoder}     & code generation LM decoder only transformer                    & \cite{Huang2023-dq}                                                                                                                                                                                                                                               \\
\textbf{llama}       & Meta open-source LLM; released 2023                            & \cite{Choi2023-hb}, \cite{Duan2023-zr}, \cite{Huang2023-dq},   \cite{Kumar2023-hl}, \cite{Kumar2023-ji}, \cite{Li2023-hf},   \cite{Lin2023-dn}, \cite{Manakul2023-yk}, \cite{Zou2023-it}                                                                          \\
\textbf{mpt}         & open source LLM from MosaicML                                  & \cite{Zou2023-it}                                                                                                                                                                                                                                                 \\
\textbf{opt}         & Open Pre-trained Transformer architecture; proposed by Meta    & \cite{Duan2023-zr}, \cite{Kuhn2023-yi}, \cite{Lin2023-dn}                                                                                                                                                                                                         \\
\textbf{palm-2}      & Google's ``next gen'' LM; 2023 release                           & \cite{Ren2023-pb}, \cite{Wu2023-my}                                                                                                                                                                                                                               \\
\textbf{roberta}     & evolution of BERT with improved training                       & \cite{Choi2023-hb}                                                                                                                                                                                                                                                \\
\textbf{santacoder}  & code generation LM trained on Github code                      & \cite{Huang2023-dq}                                                                                                                                                                                                                                               \\
\textbf{t5}          & Text-to-Text Transfer Transformer; Google                      & \cite{Choi2023-hb}                                                                                                                                                                                                                                                \\
\textbf{vicuna}      & open-source fune-tuned on LLaMA                                & \cite{Amayuelas2023-aa}, \cite{Li2023-hf}, \cite{Xiong2023-ve},   \cite{Zou2023-it}                                                                                                                                                                               \\* \bottomrule
\end{longtable}

\subsection{Referenced Datasets}\label{sec:datasets}
Varied datasets and benchmarks are necessary for evaluating the performance, and limitations, of LLMs. In the papers that we studied there were 42 datasets referenced. A list of them can be found in Table~\ref{tab:datasets}\footnote{One `dataset' listed in Table~\ref{tab:datasets} is called `general'. This represents analysis of concepts generally applicable across all datasets.}. While far from exhaustive, these datasets span several different common and relevant use-cases for LLMs. Such use cases include the more typical and general \emph{question-answering} (QA), \emph{text summarization}, and \emph{language understanding} tasks, but also include those drawn from more niche datasets such as \emph{adversarial robustness}, \emph{business ethics}, \emph{math}, \emph{science}, \emph{coding}, \emph{robotics}, \emph{music}, and \emph{law}.

Most of the listed datasets could be considered ``standard'' AI/ML datasets. There is a small subset that are more typically used in human psychology and behavioral science (i.e. `big five personality', `g7 members political', `myers-briggs', and `political-compass'). These tests were used by some researchers to assess traits and alignment normally assigned only to humans in an effort to better understand the behavior of LLMs. 

This highlights a crucial point: with LLMs having enough capability to take tests traditionally meant for humans there is need to extend evaluation from tests traditionally meant for `machines'. Previous datasets provide useful assessments in some ways, but come up lacking in others.

Finally, several `meta' or `conglomerate' datasets have started to emerge. These larger datasets include a collection of smaller datasets. These kind of benchmark datasets are useful because, while in theory it would be nice to test LLM models on \emph{every} possible use case, pragmatically the number of tests needs to be scaled with risk and focused on use-case specific vulnerabilities. Examples of such benchmarks are:

\begin{description}
    \item [`big bench':] The Beyond Imitation Game (BIG) Bench dataset includes many smaller datasets that are meant to be used to evaluate the extent to which a model has `understanding' of certain concepts and is not solely imitating understanding.
    \item [`socket`:] The Social Knowledge Evaluations Tests (SocKET) are a conglomeration of evaluations focused on quantifying the degree to which a model has `understanding' of social knowledge including topics such as empathy and humor.
\end{description}

\begin{longtable}{@{}p{0.35\textwidth}p{0.45\textwidth}p{0.2\textwidth}@{}}
\caption{Summary of datasets used in surveyed papers}
\label{tab:datasets}\\
\toprule
\multicolumn{1}{c}{\textbf{Dataset}}     & \multicolumn{1}{c}{\textbf{Description}}                                                                                     & \multicolumn{1}{c}{\textbf{Papers}}                                                                                                                                                                                                          \\* \midrule
\endfirsthead
\multicolumn{3}{c}%
{{\bfseries Table \thetable\ continued from previous page}} \\
\toprule
\multicolumn{1}{c}{\textbf{Dataset}}     & \multicolumn{1}{c}{\textbf{Description}}                                                                                     & \multicolumn{1}{c}{\textbf{Papers}}                                                                                                                                                                                                          \\* \midrule
\endhead
\bottomrule
\endfoot
\endlastfoot
\textbf{advbench}                        & dataset for adversarial robustness                                                                                           & \cite{Kumar2023-hl}, \cite{Zou2023-it}                                                                                                                                                                                                       \\
\textbf{bertology}                       & tools for accessing inner representations of BERT                                                                            & \cite{Ganguli2022-zr}                                                                                                                                                                                                                        \\
\textbf{BBQ}                             & hand build bias benchmark for QA                                                                                             & \cite{Turpin2023-kl}                                                                                                                                                                                                                         \\
\textbf{big bench}                       & Beyond the Imitation Game (BIG) collaborative benchmark for evaluating   LLMs                                                & \cite{Askell2021-mf}, \cite{Ganguli2022-zr}, \cite{Turpin2023-kl},   \cite{Wei2022-zd}                                                                                                                                                       \\
\textbf{big five personality}            & personality test proposed in 1949                                                                                            & \cite{Rutinowski2023-kk}                                                                                                                                                                                                                     \\
\textbf{biz-ethics}                      & subset of MMLU                                                                                                               & \cite{Xiong2023-ve}                                                                                                                                                                                                                          \\
\textbf{chords-db}                       & javascript database of string instrument chords                                                                              & \cite{Wu2023-my}                                                                                                                                                                                                                             \\
\textbf{cnn/daily mail}                  & text summarization dataset                                                                                                   & \cite{Huang2023-dq}                                                                                                                                                                                                                          \\
\textbf{compas}                          & inmate recidivism risk score                                                                                                 & \cite{Ganguli2022-zr}, \cite{Rutinowski2023-kk}                                                                                                                                                                                              \\
\textbf{coqa}                            & converastional question answering challenge                                                                                  & \cite{Duan2023-zr}, \cite{Kuhn2023-yi}, \cite{Lin2023-dn}                                                                                                                                                                                    \\
\textbf{custom}                          & custom-made dataset                                                                                                          & \cite{Wu2023-my}                                                                                                                                                                                                                             \\
\textbf{dark factor}                     & personality test for quantifying aversive personality traits                                                                 & \cite{Rutinowski2023-kk}                                                                                                                                                                                                                     \\
\textbf{dateund}                         & dataset designed to test LLM ability to understand dates                                                                     & \cite{Xiong2023-ve}                                                                                                                                                                                                                          \\
\textbf{eli5-category}                   & dataset for long-form question answering                                                                                     & \cite{Huang2023-dq}                                                                                                                                                                                                                          \\
\textbf{folio}                           & expert-written,   open-domain, logically complex and diverse dataset for natural   language reasoning with first-order logic & \cite{Wu2023-my}                                                                                                                                                                                                                             \\
\textbf{g7 members political}            & political affiliation tests from G7 member states                                                                            & \cite{Rutinowski2023-kk}                                                                                                                                                                                                                     \\
\textbf{general}                         & analysis/discussion of concepts generally applicable across datasets                                                         & \cite{Angelopoulos2021-uw}, \cite{Bender2021-wc},   \cite{Bommasani2021-ir}, \cite{Christ2023-bn},   \cite{OWASP_Foundation_undated-qz}, \cite{Shevlane2023-rp},   \cite{Steinhardt_undated-vi}, \cite{Weidinger2021-hs}, \cite{Wolf2023-nq} \\
\textbf{gsm8k}                           & grade school math word problems                                                                                              & \cite{Xiong2023-ve}                                                                                                                                                                                                                          \\
\textbf{hardware tabletop rearrangement} & robotic rearrangement task                                                                                                   & \cite{Ren2023-pb}                                                                                                                                                                                                                            \\
\textbf{hellaswag}                       & dataset to study grounded common-sense inference                                                                             & \cite{Askell2021-mf}                                                                                                                                                                                                                         \\
\textbf{hh}                              & helpful and harmless RLHF dataset                                                                                            & \cite{Askell2021-mf}, \cite{Li2023-hf}                                                                                                                                                                                                       \\
\textbf{humaneval}                       & programming problems to evaluate code generation                                                                             & \cite{Huang2023-dq}, \cite{Wu2023-my}                                                                                                                                                                                                        \\
\textbf{kuq}                             & Known-Unknown Questions                                                                                                      & \cite{Amayuelas2023-aa}                                                                                                                                                                                                                      \\
\textbf{lambada}                         & dataset   to evaluate the capabilities of computational models for text understanding   by means of a word prediction task   & \cite{Askell2021-mf}                                                                                                                                                                                                                         \\
\textbf{mbpp}                            & crowd-sourced python programming problems                                                                                    & \cite{Huang2023-dq}                                                                                                                                                                                                                          \\
\textbf{mmlu}                            & Massive   Multitask Language Understanding (MMLU) benchmark to measure knowledge   acquired by LLM                           & \cite{Kumar2023-ji}                                                                                                                                                                                                                          \\
\textbf{movielens}                       & movie ratings dataset                                                                                                        & \cite{Ganguli2022-zr}                                                                                                                                                                                                                        \\
\textbf{myers-briggs}                    & well-known personality test                                                                                                  & \cite{Rutinowski2023-kk}                                                                                                                                                                                                                     \\
\textbf{natural questions}               & dataset of user questions with short and long form answers                                                                   & \cite{Lin2023-dn}                                                                                                                                                                                                                            \\
\textbf{nejm case reports}               & natural language medical case reports with voting on diagnosis                                                               & \cite{Reese2023-yb}                                                                                                                                                                                                                          \\
\textbf{political compass}               & test for evaluation of political affiliation                                                                                 & \cite{Rutinowski2023-kk}                                                                                                                                                                                                                     \\
\textbf{prf-law}                         & subset of MMLU                                                                                                               & \cite{Xiong2023-ve}                                                                                                                                                                                                                          \\
\textbf{real toxicity prompts}             & dataset of sentence snippets for evaluation of toxicity                                                                      & \cite{Ganguli2022-zr}                                                                                                                                                                                                                        \\
\textbf{sciq}                            & science exam questions                                                                                                       & \cite{Duan2023-zr}                                                                                                                                                                                                                           \\
\textbf{socket}                          & dataset   for evaluating sociability of NLP models                                                                         & \cite{Choi2023-hb}                                                                                                                                                                                                                           \\
\textbf{strategyqa}                      & benchmark where required reasoning steps are implicit in the question                                                        & \cite{Xiong2023-ve}                                                                                                                                                                                                                          \\
\textbf{trivia qa}                       & reading comprehension dataset                                                                                                & \cite{Duan2023-zr}, \cite{Kuhn2023-yi}, \cite{Lin2023-dn}                                                                                                                                                                                    \\
\textbf{truthfulqa}                      & benchmark for testing whether an LM is truthful in generated responses                                                       & \cite{Wei2022-zd}                                                                                                                                                                                                                            \\
\textbf{wic}                             & word-in-context dataset; tests whether LM can identify meaning of a word                                                     & \cite{Wei2022-zd}                                                                                                                                                                                                                            \\
\textbf{wikibio}                         & dataset of biographies from Wikipedia                                                                                        & \cite{Manakul2023-yk}                                                                                                                                                                                                                        \\
\textbf{wiki-qa}                         & annotated set of question and answer paris                                                                                   & \cite{Huang2023-dq}                                                                                                                                                                                                                          \\
\textbf{wmt 2014}                        & machine translation dataset                                                                                                  & \cite{Huang2023-dq}                                                                                                                                                                                                                          \\* \bottomrule
\end{longtable}

\section{Mitigation and Detection}
Regarding LLM vulnerabilities the body of current literature presents many more problems than solutions; this indicates the phase of research that the community is currently in. New methods for handling vulnerabilities are being rapidly being proposed, and we expect this landscape to continue to change rapidly.

Mitigation of vulnerabilities/poor performance is tightly-linked with the ability to detect such problems. Generally, literature discusses methods for detection of undesirable behavior, and then assumes using that as feedback to modify the model. Many of the new detection methods rely on quantifying performance on specialized datasets (see Section~\ref{sec:datasets}) developed with the specific purpose of testing against a particular vulnerability. Beyond that, the literature reviewed discusses some nascent approaches in the following categories:

\begin{description}
    \item [Uncertainty Quantification] -- One critical capability is to be able to quantify the uncertainty an LLM has in generated responses. Standard LLMs do not include this capability `off the shelf'. There are a couple promising approaches to address this:
    \begin{itemize}
        \item Conformal Prediction (CP) -- Conformal prediction is a method for estimating the uncertainty in a model's output~\cite{Angelopoulos2021-uw}; CP is not a technology specific to LLMs, it has been used in other domains as well. The key to CP is to use a `holdout' set that is used to calibrate a `score quantile'; in a classification task the cumulative set of classes whose softmax scores are less than the score quantile are returned; the fewer classes in this set, the less uncertainty.
        \item Quantifying response variability using Semantic Entropy (see \cite{Kuhn2023-yi},\cite{Xiong2023-ve}) -- developing a measure of semantic similarity to evaluate the range of similar responses (where `similar' can be defined in different ways) produced by a LLM to the same prompt. If the LLM produces responses that have widely varying semantic similarity the uncertainty can be seen as high.
        \item Response Ranking, and LLM-Based uncertainty estimates (see \cite{Xiong2023-ve}) -- some investigation has occurred in evaluating the extent to which an LLM is capable of assessing its own uncertainty by way of prompting the LLM to include such information in its response (for example: ``provide your confidence between 0-100\% in the response''). In the case of \emph{verbalized confidence} (where the LLM is supposed to explicitly state confidence in its response), LLMs tend to be overconfident; however using other prompting strategies such as `Top-K' (ranking the top $K$ answers), or Chain-of-Thought (asking for the explicit reasoning process that lead to an answer) seems to improve the outcomes.
    \end{itemize}
    \item [Alignment] -- Methods for coercing the underlying behavior of the LLM to align with human expectations. This typically applies to auxiliary behaviors that we not directly trained originally.
    \begin{itemize}
        \item Alignment Prompts -- This involves prepending special kinds of prompts to help direct the LLM to behave `better' (i.e. avoiding bias in its response). This method is not recommended because it may encourage the model to present an `aligned facade' without actually changing its alignment \cite{Askell2021-mf}.
        \item Context Distillation -- instead of using aligning prompts (that have several drawbacks), you can fine-tune on the aligning prompt. This would avoid having to take limited prompt space with pre-specified text, and also fine-tuning actually changes the underlying behaviors of the LLM getting away from the `aligned facade` mentioned above \cite{Askell2021-mf}
        \item Preference Model Training -- It is useful to have a well-catered dataset to help create a `preference model' that can be used for reinforcement learning improvement of LLM behavior. This is called reinforcement learning from human feedback (RLHF) and is currently state-of-the-art for LLM alignment. Askell 2021~\cite{Askell2021-mf} found that using `ranked preference' for this phase is superior to `binary preference'.
        \item Preference Model Pretraining -- This approach to alignment involves training on less-tailored preference datasets (such as Stack-Exchange ranked responses); this method is attractive largely due to the increased availability of data, where fine-tuned alignment datasets typically require expensive hand-labeled datasets. This step would be followed by a fine-tuning stage later. It has been noted that binary preferences seem to work better at this stage \cite{Askell2021-mf}.
        \item RAIN (see \cite{Li2023-hf}) -- introduces ability to rewind autoregressive outputs if a specific token leads to generation of undesired output down-stream. This is a method to allow self-alignment assuming you have the capability of detecting poor outputs.
    \end{itemize}
    \item [Processes and Governance] -- Besides traditional `technical' solutions it can be just as, or more, effective to implement processes and governance structures to ensure vulnerabilities are reduced or eliminated (see \cite{Shevlane2023-rp} and \cite{Weidinger2021-hs}). As an example mentioned in \cite{Weidinger2021-hs} is that the risk of `misinformation' form LLMs is such a large-scale problem that it is unlikely that technical solutions alone will suffice; regulation, policy, and other society-level controls are necessary to effectively mitigate this problem.
\end{description}

\subsection{Persistence of Vulnerabilities}
It is important to recognize that there are fundamental limitations to the extent to which these models can actually be aligned. It has been shown that there exist prompts that can trigger any behavior that has a finite probability being exhibited~\cite{Wolf2023-nq}. This means that if behaviors aren't \emph{eliminated} during the alignment process they will not be guaranteed safe against prompt attacks. The same paper also found that using `personas' (using prompts to ask the LLM to respond as if it were `an expert' or some other non-LLM role) can serve as a shortcut to bypassing alignment training. One positive takeaway from~\cite{Wolf2023-nq} is that the better aligned the model, the longer the prompt has to be to elicit non-aligned responses; in this way `aligning prompts' (prepended text added to all prompts) were shown fairly effective especially when the overall prompt length is limited (see more in-depth discussion on alignment in the previous section).

\section{Summary and Conclusion}
The current generation of LLMs have greatly improved performance due to emergent capabilities (at least this seems the case from current thought) that have enable marked improvement an various different kinds of tasks. This increase of capability came as a surprise to many, and has led many to seriously consider adoption of LLMs to aid or enable various technologies. In the process many limitations and vulnerabilities have been highlighted or discovered. These vulnerabilities impede the possible application of LLMs in many technologies, this is especially true with respect to usage in intelligence and safety-critical use-cases where high levels of assurance are required to ensure performance lies within expected bounds.

This report highlights some of the categories of vulnerabilities, and where in the life-cycle they are likely to fall. Several mitigation strategies have also been identified. This information should be helpful in understanding the current research landscape, and guiding further research efforts.

\newpage
\printbibliography

@ARTICLE{Huang2023-dq,
  title         = "Look Before You Leap: An Exploratory Study of Uncertainty
                   Measurement for Large Language Models",
  author        = "Huang, Yuheng and Song, Jiayang and Wang, Zhijie and Chen,
                   Huaming and Ma, Lei",
  month         =  jul,
  year          =  2023,
  archivePrefix = "arXiv",
  primaryClass  = "cs.SE",
  eprint        = "2307.10236"
}

@ARTICLE{Reese2023-yb,
  title    = "On the limitations of large language models in clinical diagnosis",
  author   = "Reese, Justin T and Danis, Daniel and Caulfied, J Harry and
              Casiraghi, Elena and Valentini, Giorgio and Mungall, Christopher
              J and Robinson, Peter N",
  journal  = "medRxiv",
  month    =  jul,
  year     =  2023,
  language = "en"
}

@ARTICLE{Duan2023-zr,
  title         = "Shifting Attention to Relevance: Towards the Uncertainty
                   Estimation of Large Language Models",
  author        = "Duan, Jinhao and Cheng, Hao and Wang, Shiqi and Wang, Chenan
                   and Zavalny, Alex and Xu, Renjing and Kailkhura, Bhavya and
                   Xu, Kaidi",
  month         =  jul,
  year          =  2023,
  archivePrefix = "arXiv",
  primaryClass  = "cs.CL",
  eprint        = "2307.01379"
}

@ARTICLE{Xiong2023-ve,
  title         = "Can {LLMs} Express Their Uncertainty? An Empirical
                   Evaluation of Confidence Elicitation in {LLMs}",
  author        = "Xiong, Miao and Hu, Zhiyuan and Lu, Xinyang and Li, Yifei
                   and Fu, Jie and He, Junxian and Hooi, Bryan",
  month         =  jun,
  year          =  2023,
  archivePrefix = "arXiv",
  primaryClass  = "cs.CL",
  eprint        = "2306.13063"
}

@ARTICLE{Lin2023-dn,
  title         = "Generating with Confidence: Uncertainty Quantification for
                   Black-box Large Language Models",
  author        = "Lin, Zhen and Trivedi, Shubhendu and Sun, Jimeng",
  month         =  may,
  year          =  2023,
  archivePrefix = "arXiv",
  primaryClass  = "cs.CL",
  eprint        = "2305.19187"
}

@ARTICLE{Amayuelas2023-aa,
  title         = "Knowledge of Knowledge: Exploring {Known-Unknowns}
                   Uncertainty with Large Language Models",
  author        = "Amayuelas, Alfonso and Pan, Liangming and Chen, Wenhu and
                   Wang, William",
  month         =  may,
  year          =  2023,
  archivePrefix = "arXiv",
  primaryClass  = "cs.CL",
  eprint        = "2305.13712"
}

@ARTICLE{Choi2023-hb,
  title         = "Do {LLMs} Understand Social Knowledge? Evaluating the
                   Sociability of Large Language Models with {SocKET} Benchmark",
  author        = "Choi, Minje and Pei, Jiaxin and Kumar, Sagar and Shu, Chang
                   and Jurgens, David",
  month         =  may,
  year          =  2023,
  archivePrefix = "arXiv",
  primaryClass  = "cs.CL",
  eprint        = "2305.14938"
}

@ARTICLE{Wu2023-my,
  title         = "Reasoning or Reciting? Exploring the Capabilities and
                   Limitations of Language Models Through Counterfactual Tasks",
  author        = "Wu, Zhaofeng and Qiu, Linlu and Ross, Alexis and
                   Aky{\"u}rek, Ekin and Chen, Boyuan and Wang, Bailin and Kim,
                   Najoung and Andreas, Jacob and Kim, Yoon",
  month         =  jul,
  year          =  2023,
  archivePrefix = "arXiv",
  primaryClass  = "cs.CL",
  eprint        = "2307.02477"
}

@ARTICLE{Wolf2023-nq,
  title         = "Fundamental Limitations of Alignment in Large Language
                   Models",
  author        = "Wolf, Yotam and Wies, Noam and Avnery, Oshri and Levine,
                   Yoav and Shashua, Amnon",
  month         =  apr,
  year          =  2023,
  archivePrefix = "arXiv",
  primaryClass  = "cs.CL",
  eprint        = "2304.11082"
}

@ARTICLE{Zou2023-it,
  title         = "Universal and Transferable Adversarial Attacks on Aligned
                   Language Models",
  author        = "Zou, Andy and Wang, Zifan and Zico Kolter, J and Fredrikson,
                   Matt",
  month         =  jul,
  year          =  2023,
  archivePrefix = "arXiv",
  primaryClass  = "cs.CL",
  eprint        = "2307.15043"
}

@ARTICLE{Liu2023-yx,
  title         = "Trustworthy {LLMs}: a Survey and Guideline for Evaluating
                   Large Language Models' Alignment",
  author        = "Liu, Yang and Yao, Yuanshun and Ton, Jean-Francois and
                   Zhang, Xiaoying and Guo, Ruocheng and Cheng, Hao and
                   Klochkov, Yegor and Taufiq, Muhammad Faaiz and Li, Hang",
  month         =  aug,
  year          =  2023,
  archivePrefix = "arXiv",
  primaryClass  = "cs.AI",
  eprint        = "2308.05374"
}

@ARTICLE{Ren2023-pb,
  title         = "Robots That Ask For Help: Uncertainty Alignment for Large
                   Language Model Planners",
  author        = "Ren, Allen Z and Dixit, Anushri and Bodrova, Alexandra and
                   Singh, Sumeet and Tu, Stephen and Brown, Noah and Xu, Peng
                   and Takayama, Leila and Xia, Fei and Varley, Jake and Xu,
                   Zhenjia and Sadigh, Dorsa and Zeng, Andy and Majumdar,
                   Anirudha",
  month         =  jul,
  year          =  2023,
  archivePrefix = "arXiv",
  primaryClass  = "cs.RO",
  eprint        = "2307.01928"
}

@ARTICLE{Wei2022-zd,
  title         = "Emergent Abilities of Large Language Models",
  author        = "Wei, Jason and Tay, Yi and Bommasani, Rishi and Raffel,
                   Colin and Zoph, Barret and Borgeaud, Sebastian and Yogatama,
                   Dani and Bosma, Maarten and Zhou, Denny and Metzler, Donald
                   and Chi, Ed H and Hashimoto, Tatsunori and Vinyals, Oriol
                   and Liang, Percy and Dean, Jeff and Fedus, William",
  month         =  jun,
  year          =  2022,
  archivePrefix = "arXiv",
  primaryClass  = "cs.CL",
  eprint        = "2206.07682"
}

@ARTICLE{Kumar2023-hl,
  title         = "Certifying {LLM} Safety against Adversarial Prompting",
  author        = "Kumar, Aounon and Agarwal, Chirag and Srinivas, Suraj and
                   Feizi, Soheil and Lakkaraju, Hima",
  month         =  sep,
  year          =  2023,
  archivePrefix = "arXiv",
  primaryClass  = "cs.CL",
  eprint        = "2309.02705"
}

@ARTICLE{Ganguli2022-zr,
  title         = "Predictability and Surprise in Large Generative Models",
  author        = "Ganguli, Deep and Hernandez, Danny and Lovitt, Liane and
                   DasSarma, Nova and Henighan, Tom and Jones, Andy and Joseph,
                   Nicholas and Kernion, Jackson and Mann, Ben and Askell,
                   Amanda and Bai, Yuntao and Chen, Anna and Conerly, Tom and
                   Drain, Dawn and Elhage, Nelson and El Showk, Sheer and Fort,
                   Stanislav and Hatfield-Dodds, Zac and Johnston, Scott and
                   Kravec, Shauna and Nanda, Neel and Ndousse, Kamal and
                   Olsson, Catherine and Amodei, Daniela and Amodei, Dario and
                   Brown, Tom and Kaplan, Jared and McCandlish, Sam and Olah,
                   Chris and Clark, Jack",
  month         =  feb,
  year          =  2022,
  archivePrefix = "arXiv",
  primaryClass  = "cs.CY",
  eprint        = "2202.07785"
}

@MISC{Steinhardt_undated-vi,
  title        = "On The Risks of Emergent Behavior in Foundation Models",
  author       = "Steinhardt, Jacob",
  howpublished = "\url{https://bounded-regret.ghost.io/on-the-risks-of-emergent-behavior-in-foundation-models/}",
  note         = "Accessed: 2023-9-6"
}

@INPROCEEDINGS{Bender2021-wc,
  title     = "On the Dangers of Stochastic Parrots: Can Language Models Be Too
               Big?",
  booktitle = "Proceedings of the 2021 {ACM} Conference on Fairness,
               Accountability, and Transparency",
  author    = "Bender, Emily M and Gebru, Timnit and McMillan-Major, Angelina
               and Shmitchell, Shmargaret",
  publisher = "Association for Computing Machinery",
  pages     = "610--623",
  series    = "FAccT '21",
  month     =  mar,
  year      =  2021,
  address   = "New York, NY, USA",
  location  = "Virtual Event, Canada"
}

@ARTICLE{Bommasani2021-ir,
  title         = "On the Opportunities and Risks of Foundation Models",
  author        = "Bommasani, Rishi and Hudson, Drew A and Adeli, Ehsan and
                   Altman, Russ and Arora, Simran and von Arx, Sydney and
                   Bernstein, Michael S and Bohg, Jeannette and Bosselut,
                   Antoine and Brunskill, Emma and Brynjolfsson, Erik and Buch,
                   Shyamal and Card, Dallas and Castellon, Rodrigo and
                   Chatterji, Niladri and Chen, Annie and Creel, Kathleen and
                   Davis, Jared Quincy and Demszky, Dora and Donahue, Chris and
                   Doumbouya, Moussa and Durmus, Esin and Ermon, Stefano and
                   Etchemendy, John and Ethayarajh, Kawin and Fei-Fei, Li and
                   Finn, Chelsea and Gale, Trevor and Gillespie, Lauren and
                   Goel, Karan and Goodman, Noah and Grossman, Shelby and Guha,
                   Neel and Hashimoto, Tatsunori and Henderson, Peter and
                   Hewitt, John and Ho, Daniel E and Hong, Jenny and Hsu, Kyle
                   and Huang, Jing and Icard, Thomas and Jain, Saahil and
                   Jurafsky, Dan and Kalluri, Pratyusha and Karamcheti,
                   Siddharth and Keeling, Geoff and Khani, Fereshte and
                   Khattab, Omar and Koh, Pang Wei and Krass, Mark and Krishna,
                   Ranjay and Kuditipudi, Rohith and Kumar, Ananya and Ladhak,
                   Faisal and Lee, Mina and Lee, Tony and Leskovec, Jure and
                   Levent, Isabelle and Li, Xiang Lisa and Li, Xuechen and Ma,
                   Tengyu and Malik, Ali and Manning, Christopher D and
                   Mirchandani, Suvir and Mitchell, Eric and Munyikwa, Zanele
                   and Nair, Suraj and Narayan, Avanika and Narayanan, Deepak
                   and Newman, Ben and Nie, Allen and Niebles, Juan Carlos and
                   Nilforoshan, Hamed and Nyarko, Julian and Ogut, Giray and
                   Orr, Laurel and Papadimitriou, Isabel and Park, Joon Sung
                   and Piech, Chris and Portelance, Eva and Potts, Christopher
                   and Raghunathan, Aditi and Reich, Rob and Ren, Hongyu and
                   Rong, Frieda and Roohani, Yusuf and Ruiz, Camilo and Ryan,
                   Jack and R{\'e}, Christopher and Sadigh, Dorsa and Sagawa,
                   Shiori and Santhanam, Keshav and Shih, Andy and Srinivasan,
                   Krishnan and Tamkin, Alex and Taori, Rohan and Thomas, Armin
                   W and Tram{\`e}r, Florian and Wang, Rose E and Wang, William
                   and Wu, Bohan and Wu, Jiajun and Wu, Yuhuai and Xie, Sang
                   Michael and Yasunaga, Michihiro and You, Jiaxuan and
                   Zaharia, Matei and Zhang, Michael and Zhang, Tianyi and
                   Zhang, Xikun and Zhang, Yuhui and Zheng, Lucia and Zhou,
                   Kaitlyn and Liang, Percy",
  month         =  aug,
  year          =  2021,
  archivePrefix = "arXiv",
  primaryClass  = "cs.LG",
  eprint        = "2108.07258"
}

@ARTICLE{Kumar2023-ji,
  title         = "Conformal Prediction with Large Language Models for
                   {Multi-Choice} Question Answering",
  author        = "Kumar, Bhawesh and Lu, Charlie and Gupta, Gauri and Palepu,
                   Anil and Bellamy, David and Raskar, Ramesh and Beam, Andrew",
  month         =  may,
  year          =  2023,
  archivePrefix = "arXiv",
  primaryClass  = "cs.CL",
  eprint        = "2305.18404"
}

@ARTICLE{Askell2021-mf,
  title         = "A General Language Assistant as a Laboratory for Alignment",
  author        = "Askell, Amanda and Bai, Yuntao and Chen, Anna and Drain,
                   Dawn and Ganguli, Deep and Henighan, Tom and Jones, Andy and
                   Joseph, Nicholas and Mann, Ben and DasSarma, Nova and
                   Elhage, Nelson and Hatfield-Dodds, Zac and Hernandez, Danny
                   and Kernion, Jackson and Ndousse, Kamal and Olsson,
                   Catherine and Amodei, Dario and Brown, Tom and Clark, Jack
                   and McCandlish, Sam and Olah, Chris and Kaplan, Jared",
  month         =  dec,
  year          =  2021,
  archivePrefix = "arXiv",
  primaryClass  = "cs.CL",
  eprint        = "2112.00861"
}

@ARTICLE{Weidinger2021-hs,
  title         = "Ethical and social risks of harm from Language Models",
  author        = "Weidinger, Laura and Mellor, John and Rauh, Maribeth and
                   Griffin, Conor and Uesato, Jonathan and Huang, Po-Sen and
                   Cheng, Myra and Glaese, Mia and Balle, Borja and Kasirzadeh,
                   Atoosa and Kenton, Zac and Brown, Sasha and Hawkins, Will
                   and Stepleton, Tom and Biles, Courtney and Birhane, Abeba
                   and Haas, Julia and Rimell, Laura and Hendricks, Lisa Anne
                   and Isaac, William and Legassick, Sean and Irving, Geoffrey
                   and Gabriel, Iason",
  month         =  dec,
  year          =  2021,
  archivePrefix = "arXiv",
  primaryClass  = "cs.CL",
  eprint        = "2112.04359"
}

@ARTICLE{Manakul2023-yk,
  title         = "{SelfCheckGPT}: {Zero-Resource} {Black-Box} Hallucination
                   Detection for Generative Large Language Models",
  author        = "Manakul, Potsawee and Liusie, Adian and Gales, Mark J F",
  month         =  mar,
  year          =  2023,
  archivePrefix = "arXiv",
  primaryClass  = "cs.CL",
  eprint        = "2303.08896"
}

@ARTICLE{Rutinowski2023-kk,
  title         = "The {Self-Perception} and Political Biases of {ChatGPT}",
  author        = "Rutinowski, J{\'e}r{\^o}me and Franke, Sven and Endendyk,
                   Jan and Dormuth, Ina and Pauly, Markus",
  month         =  apr,
  year          =  2023,
  archivePrefix = "arXiv",
  primaryClass  = "cs.CY",
  eprint        = "2304.07333"
}

@ARTICLE{Turpin2023-kl,
  title         = "Language Models Don't Always Say What They Think: Unfaithful
                   Explanations in {Chain-of-Thought} Prompting",
  author        = "Turpin, Miles and Michael, Julian and Perez, Ethan and
                   Bowman, Samuel R",
  month         =  may,
  year          =  2023,
  archivePrefix = "arXiv",
  primaryClass  = "cs.CL",
  eprint        = "2305.04388"
}

@ARTICLE{Kuhn2023-yi,
  title         = "Semantic Uncertainty: Linguistic Invariances for Uncertainty
                   Estimation in Natural Language Generation",
  author        = "Kuhn, Lorenz and Gal, Yarin and Farquhar, Sebastian",
  month         =  feb,
  year          =  2023,
  archivePrefix = "arXiv",
  primaryClass  = "cs.CL",
  eprint        = "2302.09664"
}

@ARTICLE{Li2023-hf,
  title         = "{RAIN}: Your Language Models Can Align Themselves without
                   Finetuning",
  author        = "Li, Yuhui and Wei, Fangyun and Zhao, Jinjing and Zhang, Chao
                   and Zhang, Hongyang",
  month         =  sep,
  year          =  2023,
  archivePrefix = "arXiv",
  primaryClass  = "cs.CL",
  eprint        = "2309.07124"
}

@ARTICLE{Christ2023-bn,
  title         = "Undetectable Watermarks for Language Models",
  author        = "Christ, Miranda and Gunn, Sam and Zamir, Or",
  month         =  may,
  year          =  2023,
  archivePrefix = "arXiv",
  primaryClass  = "cs.CR",
  eprint        = "2306.09194"
}

@ARTICLE{Angelopoulos2021-uw,
  title         = "A gentle introduction to conformal prediction and
                   distribution-free uncertainty quantification",
  author        = "Angelopoulos, Anastasios N and Bates, Stephen",
  month         =  jul,
  year          =  2021,
  copyright     = "http://arxiv.org/licenses/nonexclusive-distrib/1.0/",
  archivePrefix = "arXiv",
  primaryClass  = "cs.LG",
  eprint        = "2107.07511"
}

@ARTICLE{Shevlane2023-rp,
  title         = "Model evaluation for extreme risks",
  author        = "Shevlane, Toby and Farquhar, Sebastian and Garfinkel, Ben
                   and Phuong, Mary and Whittlestone, Jess and Leung, Jade and
                   Kokotajlo, Daniel and Marchal, Nahema and Anderljung, Markus
                   and Kolt, Noam and Ho, Lewis and Siddarth, Divya and Avin,
                   Shahar and Hawkins, Will and Kim, Been and Gabriel, Iason
                   and Bolina, Vijay and Clark, Jack and Bengio, Yoshua and
                   Christiano, Paul and Dafoe, Allan",
  month         =  may,
  year          =  2023,
  archivePrefix = "arXiv",
  primaryClass  = "cs.AI",
  eprint        = "2305.15324"
}

@MISC{OWASP_Foundation_undated-qz,
  title       = "Top 10 for large language model applications: {OWASP}
                 Foundation Web Respository",
  author      = "{OWASP Foundation}",
  institution = "Github",
  language    = "en"
}

\end{document}